\documentclass[conference,letterpaper]{IEEEtran}
\IEEEoverridecommandlockouts

\usepackage{url} 
\usepackage{fancyhdr}
\usepackage[dvipdf]{graphicx}
\usepackage{indent}
\usepackage{amsmath,amssymb}
\usepackage{paralist}
\usepackage{eclbkbox}
\usepackage{subfigure}

\makeatletter
\def\tbcaption{\def\@captype{table}\caption}
\def\figcaption{\def\@captype{figure}\caption}

\newcommand{\bvec}[1]{\mbox{\boldmath $#1$}}

\begin{document}
\title{A Generation Method of Immunological Memory \\in Clonal Selection Algorithm by using \\Restricted Boltzmann Machines
\thanks{\copyright 2015 IEEE. Personal use of this material is permitted. Permission from IEEE must be obtained for all other uses, in any current or future media, including reprinting/republishing this material for advertising or promotional purposes, creating new collective works, for resale or redistribution to servers or lists, or reuse of any copyrighted component of this work in other works.}}

\author{
\IEEEauthorblockN{Shin Kamada}
\IEEEauthorblockA{Dept. of Intelligent Systems, \\
Graduate School of Information Sciences, \\
Hiroshima City University\\
3-4-1, Ozuka-Higashi, Asa-Minami-ku,\\
Hiroshima, 731-3194, Japan\\
Email: da65002@e.hiroshima-cu.ac.jp}
\and
\IEEEauthorblockN{Takumi Ichimura}
\IEEEauthorblockA{Faculty of Management and Information Systems,\\
Prefectural University of Hiroshima\\\\
1-1-71, Ujina-Higashi, Minami-ku,\\
Hiroshima, 734-8559, Japan\\
Email: ichimura@pu-hiroshima.ac.jp}
}

\maketitle

\fancypagestyle{plain}{
\fancyhf{}	
\fancyfoot[L]{}
\fancyfoot[C]{}
\fancyfoot[R]{}
\renewcommand{\headrulewidth}{0pt}
\renewcommand{\footrulewidth}{0pt}
}

\pagestyle{fancy}{
\fancyhf{}
\fancyfoot[R]{}}
\renewcommand{\headrulewidth}{0pt}
\renewcommand{\footrulewidth}{0pt}

\begin{abstract}
Recently, a high technique of image processing is required to extract the image features in real time. In our research, the tourist subject data are collected from the Mobile Phone based Participatory Sensing (MPPS) system. Each record consists of image files with GPS, geographic location name, user's numerical evaluation, and comments written in natural language at sightseeing spots where a user really visits. In our previous research, the famous landmarks in sightseeing spot can be detected by Clonal Selection Algorithm with Immunological Memory Cell (CSAIM). However, some landmarks was not detected correctly by the previous method because they didn't have enough amount of information for the feature extraction. In order to improve the weakness, we propose the generation method of immunological memory by Restricted Boltzmann Machines. To verify the effectiveness of the method, some experiments for classification of the subjective data are executed by using machine learning tools for Deep Learning.
\end{abstract}

\begin{IEEEkeywords}
Image Analysis, Clonal Selection Algorithm, Immunological Memory Cells, Restricted Boltzmann Machines, Deep Learning, Smartphone based Participatory Sensing System, Knowledge Discovery
\end{IEEEkeywords}

\IEEEpubidadjcol

\section{Introduction}
\label{sec:Introduction}
The current information technology can collect various data sets because the recent tremendous technical advances in processing power, storage capacity, and network connected cloud computing. The sample record in such data set includes not only numerical values but also comments, numerical evaluation, and binary data such as pictures. The technical methods to discover knowledge in such databases are known to be a field of data mining and developed in various research fields.

Moreover, Mobile Phone based Participatory Sensing (MPPS) system involves a community of users sending personal information and participating in autonomous sensing through their mobile phones \cite{Lane2010}. Sensed data can be obtained from sensing devices such as audio, video, and motion sensors in the current smartphone. And then their data can also be obtained from external sensing devices that can communicate wirelessly to the phone. Participation of mobile phone users in sensorial data collection both from the individual and from the surrounding environment presents a wide range of opportunities for truly pervasive applications. 

In our research, the tourist subject data are collected from the Android smartphone application \cite{Android_Market}. The collected data consists of image files with GPS, geographic location name, user's numerical evaluation, and comments written in natural language at sightseeing spots where a user really visits. More than 800 subjective data in Hiroshima area are recorded in the database through this system. We have already proposed the classification method from the collected subjective data by the interactive GHSOM \cite{Ichimura12a} and the knowledge is extracted from the classification results of the interactive GHSOM by C4.5 \cite{Ichimura12b}. Moreover, the famous landmarks included in the image data of sightseeing spot were detected by Clonal Selection Algorithm with Immunological Memory Cell (CSAIM) \cite{Ichimura12c}.

CSAIM is one of the artificial immune systems which realizes the process of immune response. In CSAIM, the clustering of the antibodies generated by somatic hypermutation (HM) and receptor editing (RE) can be realized according to the training samples in data set. Each cluster of antibodies is trained to extract the features of some sample patterns by perceptron and then the representative antibody in the cluster is recorded as a memory cell after training. The classification capability of this method was high to the medical database. \cite{Ichimura14}. However, the method didn't show good performance in the experiments of image data set because it was too large amount of information to realize the extraction of the specified characteristics\cite{Ichimura13}. In order to improve this problem, we proposed the feature extraction method of CSAIM by using RBM which generates memory cells to extract the characteristic pattern from images by Deep Learning.

Deep Learning method attracts a lot of attention in the research of machine learning, especially image processing \cite{Bengio09}. The method has an advantage of not only the network structure with multi-layer but also the pre-training which is an unsupervised learning for each layer before actual training. Restricted Boltzmann Machine (RBM) \cite{Hinton12} is often used for one of pre-training methods, it can represent an probability distribution of input data set. Moreover, some machine learning tools for Deep Learning such as ``Pylearn2 \cite{Goodfellow13}'' are shared as a open source software and the techniques of image processing are required to extract the image features in real time.

The remainder of this paper is organized as follows. In Section \ref{sec:CSAIM}, the our previous proposed clonal selection algorithm with immunological memory will be explained briefly. Section \ref{sec:RBMs} describes the algorithm of RBM and Section \ref{sec:Pylearn2} explains the usage of Pylearn2 as one of machine learning tools for Deep Learning. In Section \ref{sec:Generation-RBMs}, we propose a generation method of immunological memory cells in clonal selection algorithm by using RBM. Section \ref{sec:ExperimentalResult} describes some experimental results. In Section \ref{sec:Conclusion}, we give some discussions to conclude this paper.

\section{Clonal Selection Algorithm with Immunological Memory}
\label{sec:CSAIM}
This section describes our proposed AIS model called Clonal Selection Algorithm with Immunological Memory (CSAIM) \cite{Ichimura12c, Ichimura14}.

The area of artificial immune system (AIS) has been an ever-increasing interested in not only theoretical works but applications in pattern recognition, network security, and optimization \cite{Castro1}, \cite{Dasgupta}, \cite{Perelson}. AIS uses ideas gleaned from immunology in order to develop adaptive systems capable of performing a wide range of tasks in various research areas. 

In our research, we proposed Clonal Selection Algorithm with Immunological Memory (CSAIM) \cite{Ichimura12c, Ichimura14}. The method is based on RECSA model proposed by Gao \cite{Gao}. He indicated the complementary roles of somatic hypermutation (HM) and receptor editing (RE). However, RECSA model doesn't have a mechanism which explains immune response. In CSAIM model, the generated antibodies by HM and RE can be recorded into memory cells.

\begin{figure}[tbp]
\begin{center}
\includegraphics[scale=0.5]{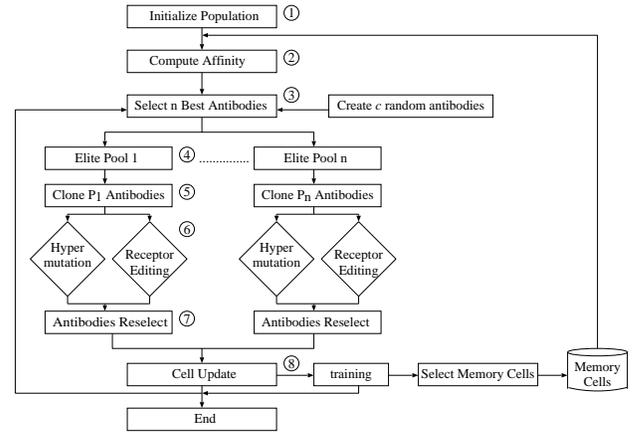}
\vspace{-3mm}
\caption{A flow of CSAIM model}
\label{fig:CSAIMmodel}
\end{center}
\end{figure}

Fig. \ref{fig:CSAIMmodel} shows a flow of CSAIM model. The process from \textcircled{1} to \textcircled{8} in Fig. \ref{fig:CSAIMmodel} is the same process as RECSA model. In CSAIM model, generated antibodies by RECSA model are classified into some categories according to their features.

\begin{figure}[tbp]
\begin{center}
\includegraphics[scale=0.55]{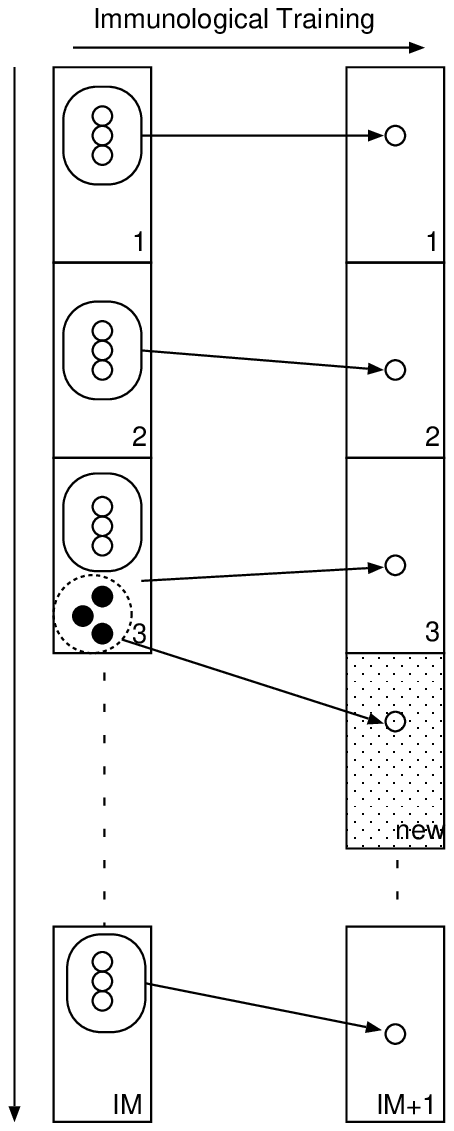}
\vspace{-3mm}
\caption{A generation method of new memory cell}
\label{fig:CSAIMmodel_clustering}
\end{center}
\end{figure}

Clustering Memory Cells are required to classify the antibodies responding the specified samples in the data set. This articles\cite{Ichimura12c, Ichimura14} realize the clustering by allocating the generated antibodies by RECSA model into some categories. The initial number of categories is predefined and a new category is created according to training situation. Fig. \ref{fig:CSAIMmodel_clustering} shows the generation method of new memory cell. Similar antibodies crowd around an appropriate point in each category, and then only central antibody of the crowd can become a memory cell. However, we may meet that memory cells can not recognize some of samples in the data set. The real database which includes some contradictory cases shows different output patterns for the some input pattern. In such cases, some new generated antibodies by RECSA model tries to respond to the mis-classification of the samples, if the similar antibodies make a crowd.

To find the crowd of similar cases, the system checks whether the Euclidean distance between the training sample and its corresponding antibody is smaller than the predetermined parameter $\mu_{\theta}$. Let $\vec{d}=(d_1,\cdots,d_i,\cdots,d_k)$ be the elements of input signal and $\vec{h}=(h_1,\cdots,h_i,\cdots,h_k)$ be the element of antibody.

In order to calculate the distance between the sample and the antibody, the range of sample is changed to that of antibody as follows.
\vspace{-3mm}
\begin{equation}
\nonumber d_{i}^{'}=d_{i} \times \frac{h_{j}}{d_{j}} \ (d_{i} \neq  0 \wedge h_{i} \neq 0)
\end{equation}
where $d_{j}$ is the minimum value of element in the input sample.

Then, if the Euclidean distance between $\vec{d^{'}}$ and $\vec{h}$ is smaller than $\mu_{\theta}$, the antibody can respond the sample. 

Also, each of classified antibodies can be trained to learn the sample pattern by perceptron. Fig. \ref{fig:CSAIM} shows the procedure of the learning algorithm of perceptron. After the training, if training error is less than the expected value, the antibody is recorded as a new memory cell.

\vspace{-1.5mm}
\begin{center}
\begin{indentation}{0.5cm}{0.0cm}
\begin{breakbox}
\smallskip
\begin{enumerate}
\setlength{\itemsep}{-6.0mm}
\item For each antibody, calculate the output of the specified samples by using the following equation.
\vspace{-3mm}
\begin{equation}
O=\sum_{i=1}^{k} w_{i} x_{i}
\end{equation}
where $w_{i}$ is a weight for each antibody and $x_i$ is a element of input.
\label{CSAIM_output}
\item Calculate the difference $\delta$ by using output value and $\theta_q$.
\begin{equation}
 E = \frac{1}{2}\delta^{2} = \frac{1}{2}(\theta_{q} -O)^{2}
\end{equation}
where $q(1\le q \le IM)$ means the cluster of memory cells.
\item Update weights.
\begin{equation}
 w_{i}=w_{i}+\eta \delta x_{i}
\end{equation}
where $\eta$ takes a real value in $[0.1, 1.0]$.
\label{CSAIM_update}
\item The procedure from \ref{CSAIM_output}) to \ref{CSAIM_update}) is executed for the specified samples till the given number of iterations reaches or error becomes less than the expected value. The good antibodies after training are recorded in the memory.
\end{enumerate}
\end{breakbox}
\end{indentation}
\figcaption{The learning algorithm by perceptron in CSAIM}
\label{fig:CSAIM}
\end{center}

\section{Restricted Boltzmann Machines}
\label{sec:RBMs}
\begin{figure}[tbp]
\begin{center}
\includegraphics[scale=0.65]{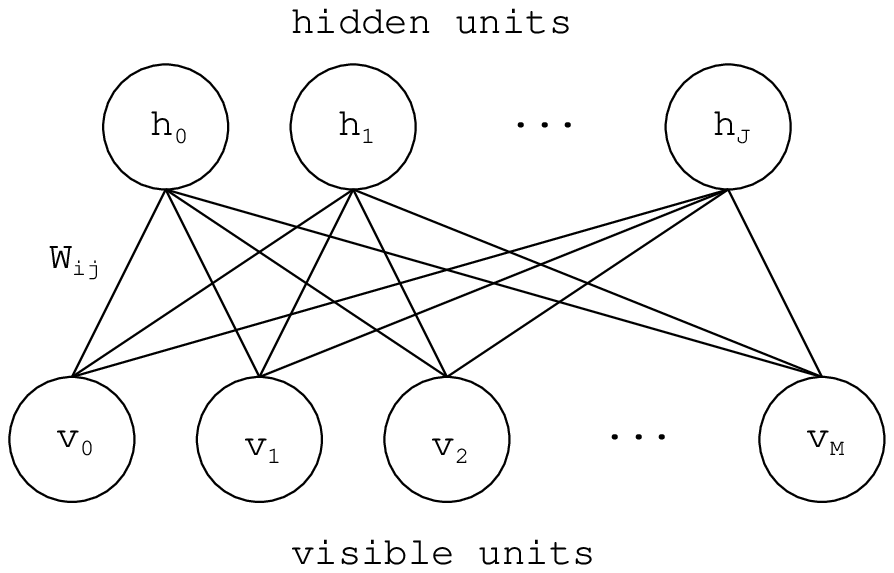}
\vspace{-3mm}
\caption{The structure of RBM}
\label{fig:rbm}
\end{center}
\end{figure}

This section explains Restricted Boltzmann Machine (RBM) \cite{Hinton12}. RBM is a generative stochastic artificial neural network and energy-based model for unsupervised learning. It consists of two layers; one is visible layer, the other is hidden layer as shown in Fig. \ref{fig:rbm}. In Boltzmann Machines network graph, each neuron is connected to all the neurons in all layers. However, because of computational complexity and stochastically independent, there are no connections between neurons in the same layer in RBM. Recently, RBM can be used as a method of pre-training in Deep Learning for fine tuning of initial parameters.

The purpose of the training of RBM is the learning of a probability distribution over the input set. Let $v_i$ and $h_j$ be the binary variables of the visible unit and hidden unit, respectively.
An energy function and probabilities of vector $\bvec{v}=\{v_0, \cdots, v_i, \cdots, v_M \}$ for visible units and $\bvec{h}=\{h_0, \cdots, h_j, \cdots, h_J \}$ for hidden units, respectively are defined as follows.

\vspace{-3mm}
\begin{equation}
\label{eq:energy}
E(\bvec{v}, \bvec{h}) = \sum_{i} b_i v_i - \sum_j c_j h_j - \sum_{i} \sum_{j} v_i W_{ij} h_j ,
\end{equation}

\vspace{-3mm}
\begin{equation}
\label{eq:prob}
p(\bvec{v}, \bvec{h})=\frac{1}{Z} \exp(-E(\bvec{v}, \bvec{h})) ,
\end{equation}

\vspace{-3mm}
\begin{equation}
\label{eq:PartitionFunction}
Z = \sum_{\bvec{v}} \sum_{\bvec{h}} \exp(-E(\bvec{v}, \bvec{h})) ,
\end{equation}

where $b_i$ is the bias for $v_i$, $c_j$ is the bias for $h_j$, and $W_{ij}$ is the weight between $v_i$ and $h_j$. $Z$ is the partition function which is given by summing over all possible pairs of visible and hidden vectors. 

The parameters of RBM are calculated by maximum likelihood estimation for the $p(\bvec{v}) = \sum_{\bvec{h}} p(\bvec{v}, \bvec{h})$ which is marginal distribution of $\bvec{v}$. However, computational complexity increases exponentially because of the calculation of all possible pairs of visible and hidden vectors during the calculation of maximum likelihood estimation. Usually, the contrastive divergence learning procedure which is a much faster algorithm of Gibbs sampling based on Markov chain Monte Carlo methods can be often used as one of the learning methods of RBM \cite{Hinton02,Tileman08}. Fig. \ref{fig:RBMs-algorithm} shows the update procedure in RBM with single-step contrastive divergence (CD-1).

\vspace{-1.5mm}
\begin{center}
\begin{indentation}{0.4cm}{-0.2cm}
\begin{breakbox}
\smallskip
\begin{enumerate}
\setlength{\itemsep}{-3.0mm}
\item Take a training sample $\bvec{v}$ from data set in visible layer.
\label{RBMs_start}
\item For each hidden unit, compute the conditional probabilitiy of $h_j$ given $\bvec{v}$ and binary visible $h_j \in \{ 0, 1 \}$.
\begin{equation}
p(h_j = 1 | \bvec{v})= {\rm sigm}(c_j + \sum_{i}W_{ij} v_i)
\end{equation}
where ${\rm sigm}()$ is a function which output a real value in $[0, 1]$ such as sigmoid function.
\item For each visible unit, compute the conditional probability of $v^{'}_i$ given $\bvec{h}$ calculated by the step 2) and binary visible $v_i^{'} \in \{ 0, 1 \}$.
\begin{equation}
p(v^{'}_i = 1 | \bvec{h})= {\rm sigm}(b_i + \sum_{j}W_{ij} h_j)
\end{equation}
\item For each hidden unit, compute the conditional probability of $h^{'}_j$ given $\bvec{v}^{'}$ calculated by the step 3) and binary visible $h_j^{'} \in \{ 0, 1 \}$.
\begin{equation}
p(h^{'}_j = 1 | \bvec{v}^{'})= {\rm sigm}(c_j + \sum_{i}W_{ij} v^{'}_i)
\end{equation}
\item Update bias and weights as follows.
\label{RBMs_end}
\begin{eqnarray}
W_{ij} &=& W_{ij} + \eta (v_i p(h_j = 1 | \bvec{v}) - v^{'}_i p(h^{'}_j = 1 | \bvec{v}^{'})) \nonumber \\
b_{i} &=& b_i + \eta (v_i - v^{'}_i) \nonumber \\
c_{j} &=& c_j + \eta (p(h_j = 1 | \bvec{v})  - p(h^{'}_j = 1 | \bvec{v}^{'}))
\end{eqnarray}
where $\eta$ is a learning rate $[0,1]$.
\item The procedure from \ref{RBMs_start}) to \ref{RBMs_end}) is executed until the terminate conditions are satisfied.
\end{enumerate}
\end{breakbox}
\end{indentation}
\figcaption{The learning algorithm in RBMs by CD-1}
\label{fig:RBMs-algorithm}
\end{center}

\section{Machine Learning Tool}
\label{sec:Pylearn2}
Pylearn2 \cite{Goodfellow13} is one of machine learning tools with libraries for Deep Learning \cite{Bengio09}. The programming code is written by python and based on Theano \cite{theano} which is numerical computation library. Pylearn2 can work well in the GPU calculation environment and it is suitable for the implementation of mathematical expressions. Pylearn2 can represent the computation result by RBM, DBN (Deep Belief Networks) \cite{Hinton06}, and other Deep Learning method visually.

Pylearn2 is shared in GitHub as an open source software \cite{pylearn2-github}. The following procedures are the instruction of the installation of Pylearn2 in Linux. 

\begin{enumerate}
\item Before the installation of Pylearn2, the RPM packages such as python-pip, python-numpy, python-scipy, python-setuptools, python-matplotlib, and Theano are required in the Linux with Red Hat Linux package management system.

\item Pylearn2 system is obtained from the Internet by using Git system as follows.

{\small \begin{verbatim}
$ git clone git://github.com/lisa-lab/
pylearn2.git
\end{verbatim} }

The software will be compiled by python and installed as follows.

{\small \begin{verbatim}
$ python setup.py build
$ sudo python setup.py install
\end{verbatim} }

\item We should define some environment variables in the Linux shell in order to use the data files and script files without absolute path.

{\small \begin{verbatim}
$ export PYLEARN2_VIEWER_COMMAND=
"eog --new-instance"
$ export PATH=$PATH:~/pylearn2/pylearn2/
scripts
\end{verbatim} }

\end{enumerate}

To use some Pylearn2 scripts, the following procedures are required.

\begin{figure}[tbp]
\begin{center}
\includegraphics[scale=0.8]{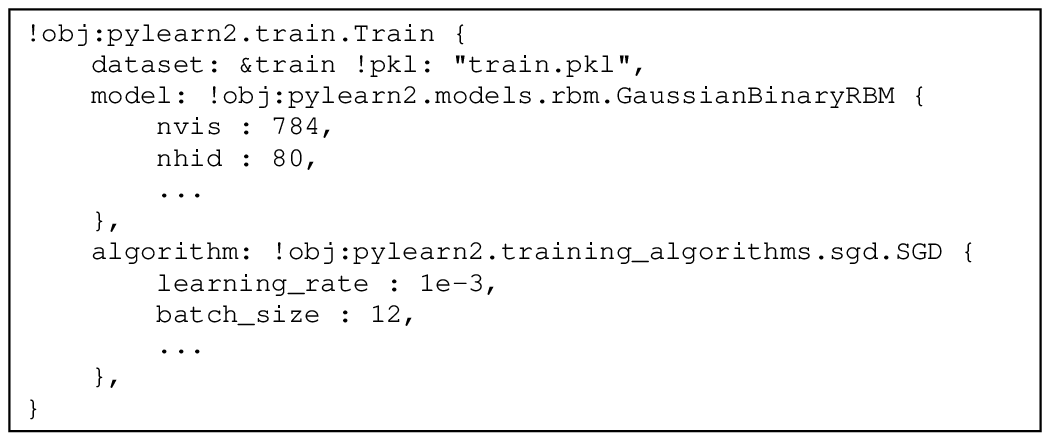}
\vspace{-3mm}
\caption{An example of YAML file}
\label{fig:yaml-example}
\end{center}
\end{figure}

\begin{enumerate}
\renewcommand{\labelenumi}{\alph{enumi})}
\item Create the data set such as CSV file or PKL file which is python serial object.

\item Write YAML file. The YAML file is a series of 2 or more XML files. Fig. \ref{fig:yaml-example} is a very simple example of YAML file. The file mainly consists of three sections: data set, model, and algorithm. In Fig. \ref{fig:yaml-example}, the input data set is ``train.pkl'' (line 2), the model is Gaussian Binary RBM (line 3-7), the algorithm is Stochastic Gradient Descent (line 8-12). We can define the training parameters for each section such as the number of visible units, learning rate, and so on. 

\item Import YAML file and run the process as follows.

{\small \begin{verbatim}
$ train.py example.yaml
\end{verbatim} }

\item After the training, we can see the training result for each iteration as follows.

{\small \begin{verbatim}
$ plot_moniter.py example.pkl
\end{verbatim} }

The file ``example.pkl'' is the output file by training. The visualization of learning weights can be appear by using the following command.

{\small \begin{verbatim}
$ show_weight.py example.pkl
\end{verbatim} }

\end{enumerate}

As mentioned above, Pylearn2 supports General-purpose computing on graphics processing units (GPGPU). An architecture for GPGPU such as Compute Unified Device Architecture (CUDA) \cite{cuda} should be required. CUDA has been developed by NVIDIA and it can give an developer direct access to the virtual instruction set and memory of the parallel computational elements. CUDA Ver 7.0 has been released at the time of writting this paper, and it is available for Windows, Mac, and Linux. 

After CUDA installation, we can set the following environment variables for Theano with Pylearn2 and run on GPU.

{\small \begin{verbatim}
$ THEANO_FLAGS=mode=FAST_RUN,device=gpu,
floatX=float32 train.py example.yaml
\end{verbatim} }

The global setting in ``.theanorc'' file is configured as follows.

{\small \begin{verbatim}
[global] 
device=gpu
floatX=float32
\end{verbatim} }

\section{A Generation method of memory cell of RBM}
\label{sec:Generation-RBMs}
\begin{figure}[tbp]
\begin{center}
\subfigure[Torii in Miyajima 1]{\includegraphics[scale=0.5]{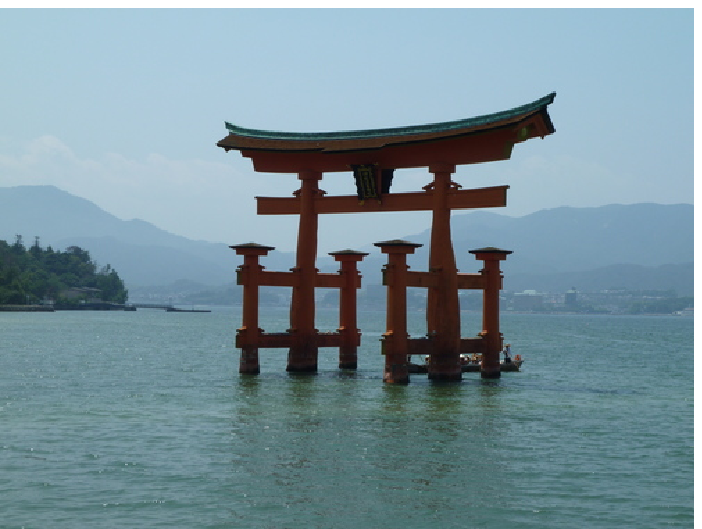}\label{fig:miyajima1}}
\subfigure[Torii in Miyajima 2]{\includegraphics[scale=0.2]{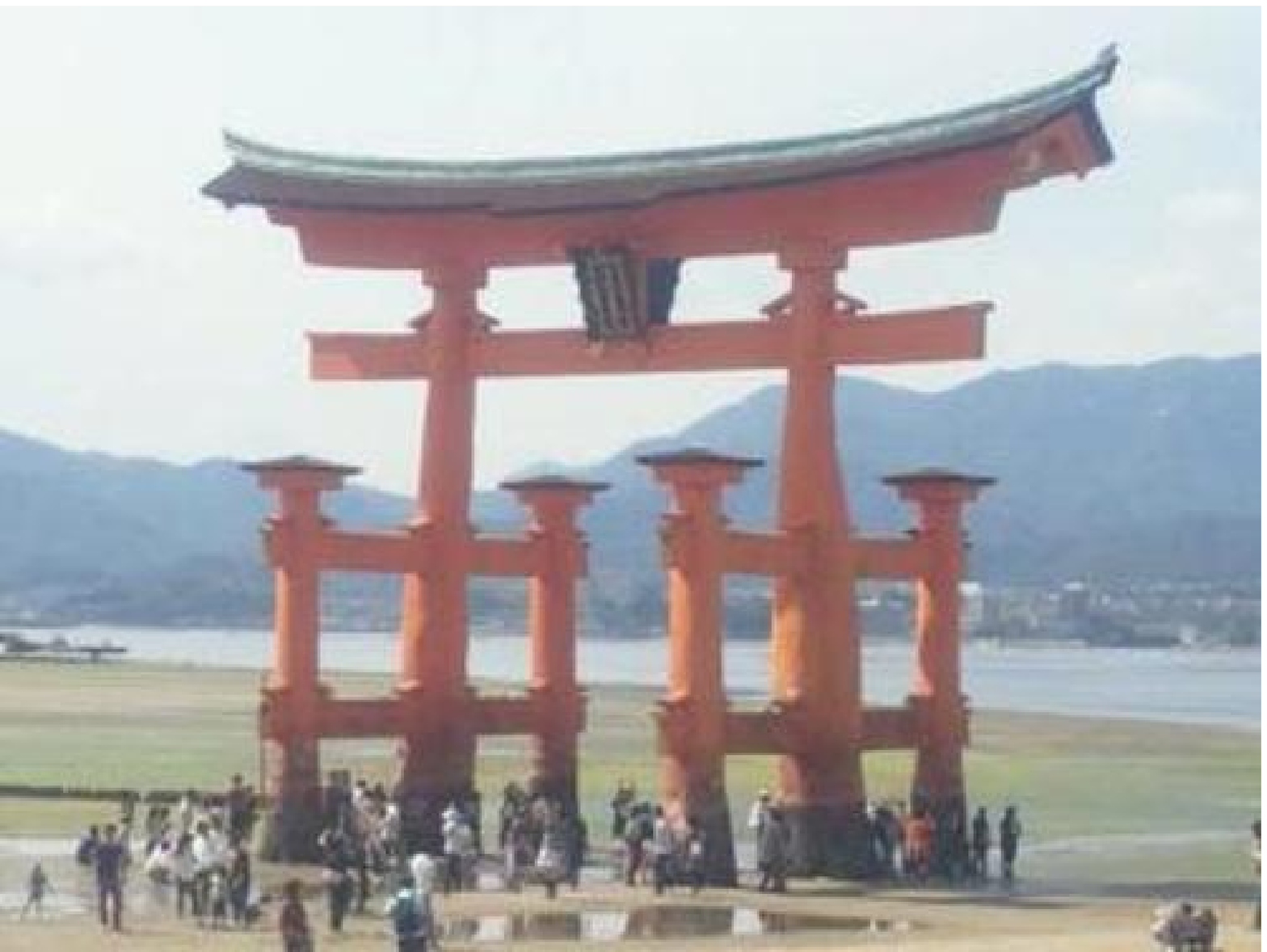}\label{fig:miyajima2}}
\subfigure[Atomic Bomb Dome 1]{\includegraphics[scale=0.5]{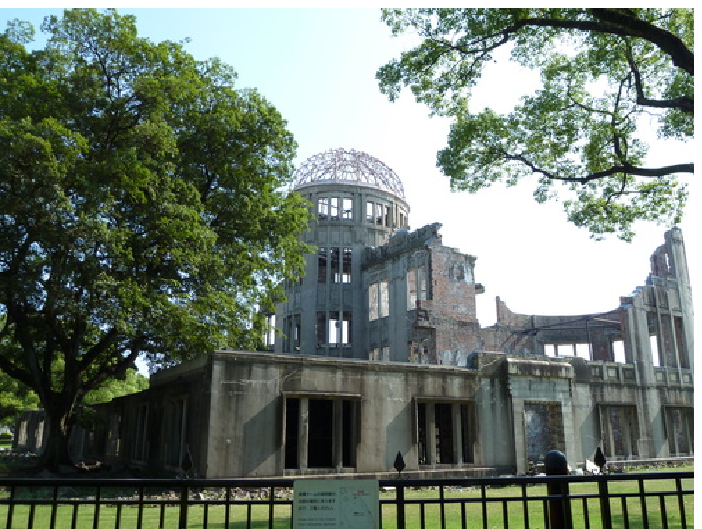}\label{fig:genbaku1}}
\subfigure[Atomic Bomb Dome 2]{\includegraphics[scale=0.5]{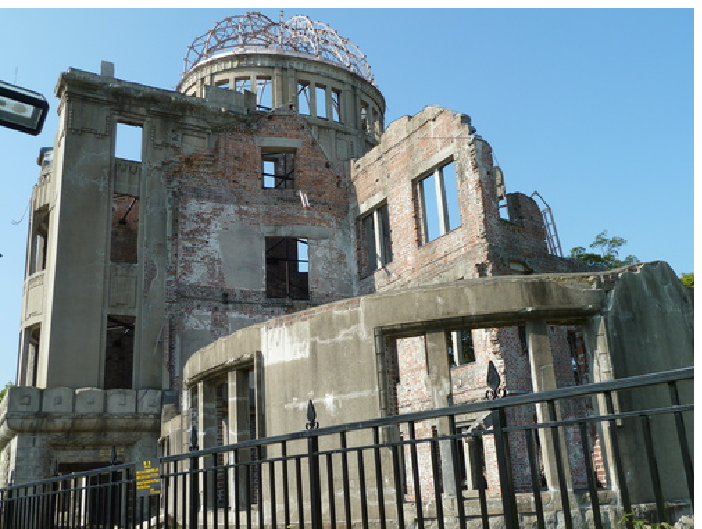}\label{fig:genbaku2}}
\subfigure[Battleship Yamato 1]{\includegraphics[scale=0.5]{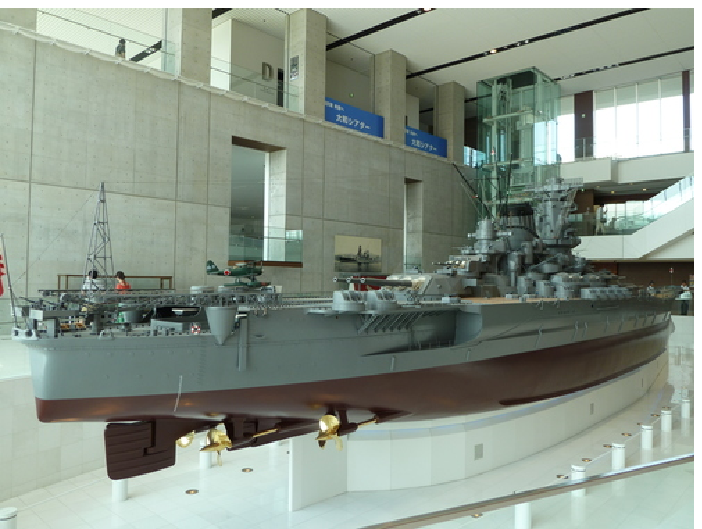}\label{fig:yamato1}}
\subfigure[Battleship Yamato 2]{\includegraphics[scale=0.5]{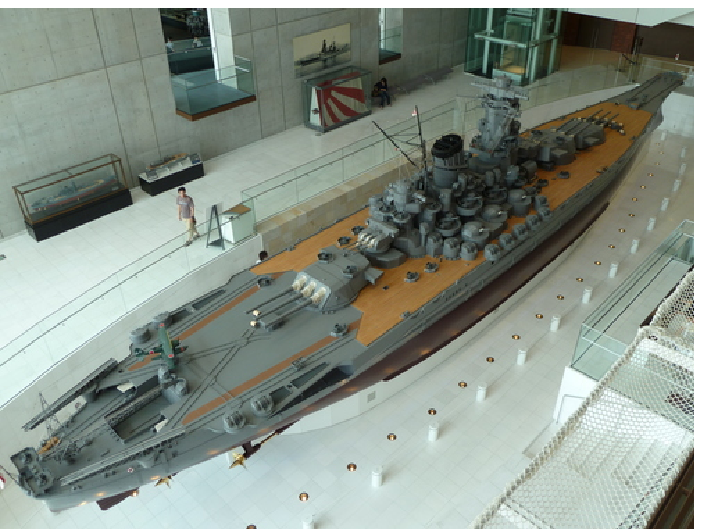}\label{fig:yamato2}}
\end{center}
\vspace{-3mm}
\caption{An example of data set}
\label{fig:dataset}
\end{figure}

As mentioned in the section \ref{sec:CSAIM}, when CSAIM model classifies the generated antibodies into some categories, the system generates memory cells responding the specified samples by perceptron. The method can provide the capability of the learning for simple patterns. However, some data which is ambiguous such as image may not be classified correctly because CSAIM model has a problem related to the number of memory cells. CSAIM model generates a new memory cell according to a new classified category as shown in Fig. \ref{fig:CSAIMmodel_clustering}. If the distinctive patterns are appeared in the data set, the system allocates a huge number of memory cells. As a result, the computation time to classify the sample case will be larger. In order to improve the weakness, we proposed the feature extraction method of sequence of pixels from a photo image to reduce the number of pixels for each memory cell \cite{Ichimura13}. Fig. \ref{fig:dataset} shows some experiments of data sets for classification of the images by using CSAIM. 
As a result, Fig. \ref{fig:miyajima1} and Fig. \ref{fig:miyajima2} were classified correctly, but Fig. \ref{fig:genbaku1} - Fig. \ref{fig:yamato2} were not well classified because the sample image has too ambiguous to extract their features compared with Fig. \ref{fig:miyajima1} and Fig. \ref{fig:miyajima2}. In this paper, we propose another CSAIM model which generates memory cells to extract the characteristic patterns from images by using RBM.

\begin{figure}[tbp]
\begin{center}
\includegraphics[scale=0.6]{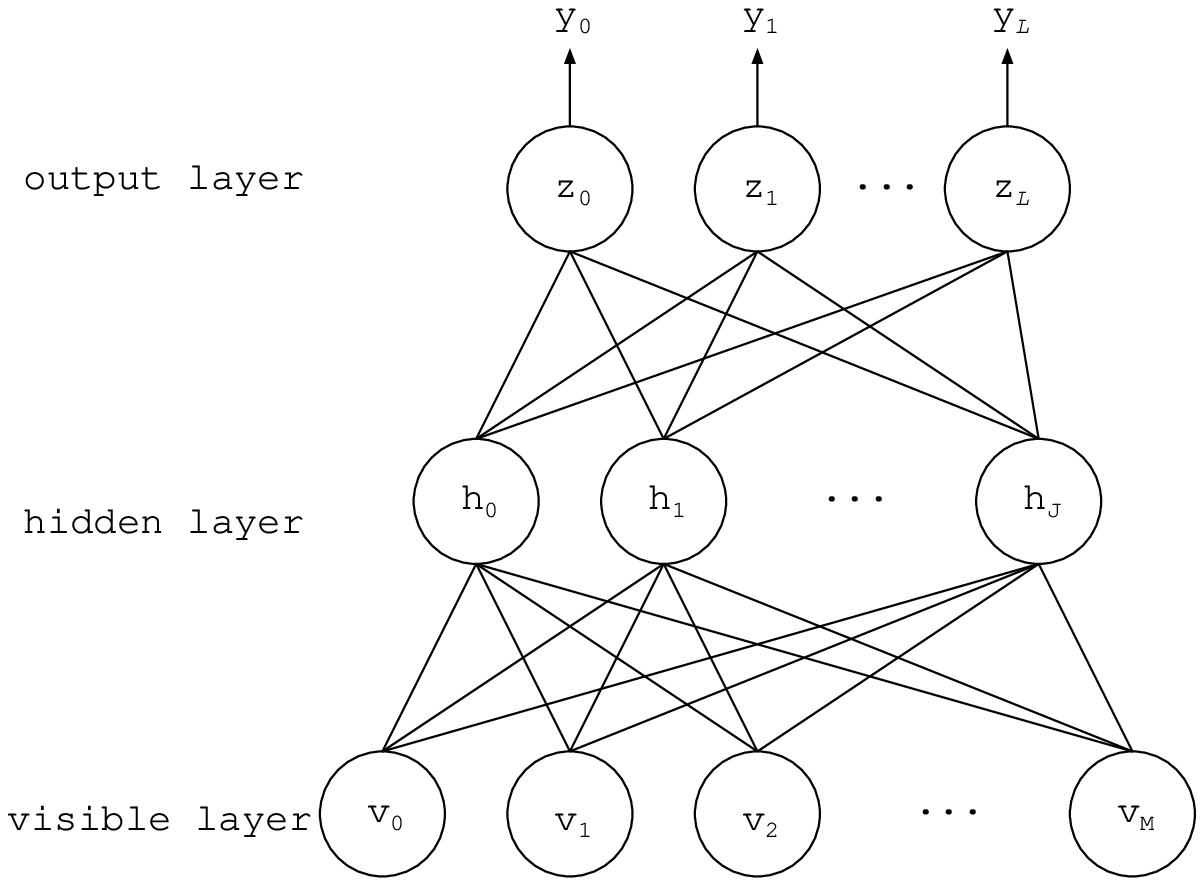}
\vspace{-3mm}
\caption{The structure of DBN}
\label{fig:dbn}
\end{center}
\end{figure}

In our proposed method, the specified samples are trained by RBM. An output layer is added into hidden layer of RBM, the weights between hidden layer and output layer are fine tuned by perceptron in order to estimate the output value for a sample pattern. Fig. \ref{fig:dbn} shows the structure of our proposed model. $\bvec{z}=\{z_0, \cdots, z_k, \cdots, h_L \}$ is a vector of output layer, $y_k$ is used to estimate the label by Softmax function as shown in Eq. (\ref{eq:SoftMax}).

\vspace{-3mm}
\begin{equation}
\label{eq:SoftMax}
y_k = \frac{\exp(z_k)}{\sum_{i}^{L} \exp(z_i)}
\end{equation}

\section{Experimental Result}
\label{sec:ExperimentalResult}

\subsection{Data Set}
To verify the effectiveness of our proposed model, we used the image data set collected from smartphone Application called ``Hiroshima Tourist Map \cite{Android_Market}''. Hiroshima Tourist Map can collect the tourist subjective data in the research field of MPPS. The collected subjective data consists of jpeg files with GPS, geographic location name, user's numerical evaluation, and comments written in natural language at sightseeing spots where a user really visits. More than 800 subjective data are recorded in the database through this system.

In this experiment, 3 kinds of images were used as a landmark spot of world heritage in Hiroshima; ``Torii'' in Miyajima, ``Atomic Bomb Dome'', and ``Battleship Yamato'' as shown in Fig. \ref{fig:dataset}. We used 8 training samples and 2 test samples for each spot, and each image is 48 $\times$ 48 pixels.

\subsection{Results}
We examined the classification capability of our proposed method. In this experiment, the following parameters were used for CSAIM: $G_{max}=100$, $m=150$, $n=100$, $Q=50$, $HM:RE=1:1$, $-1 \leq r_{w} \leq 1$, $-1 \leq r_{\theta} \leq 1$, $E_{sim}=0.05$, $t=10$, $c=10$, $\eta=0.1$, $\mu_{\theta}=0.3$, $t_{IM}=50$, $E_{min}=0.001$,  $c_{max}^{memory}=\frac{1}{2}n$, $IM =10$ (for the details of parameter description, see \cite{Ichimura12c,Ichimura14}). Moreover, the following parameters were used for RBM: model = Gaussian Binary RBM, number of visible units = 2304, number of hidden units = 80, training\_algorithms = Stochastic Gradient Descent (SGD), batch\_size = 6 in this paper. We used the computer with the following specifications: CPU = Intel(R) Core(TM) i7-2600 CPU @ 3.40GHz, GPU = GeForce GTX 750 Ti, Memory = 16GB, OS = Ubuntu 14.04.2 LTS x86\_64.

\begin{figure}[tbp]
\begin{center}
\includegraphics[scale=0.55]{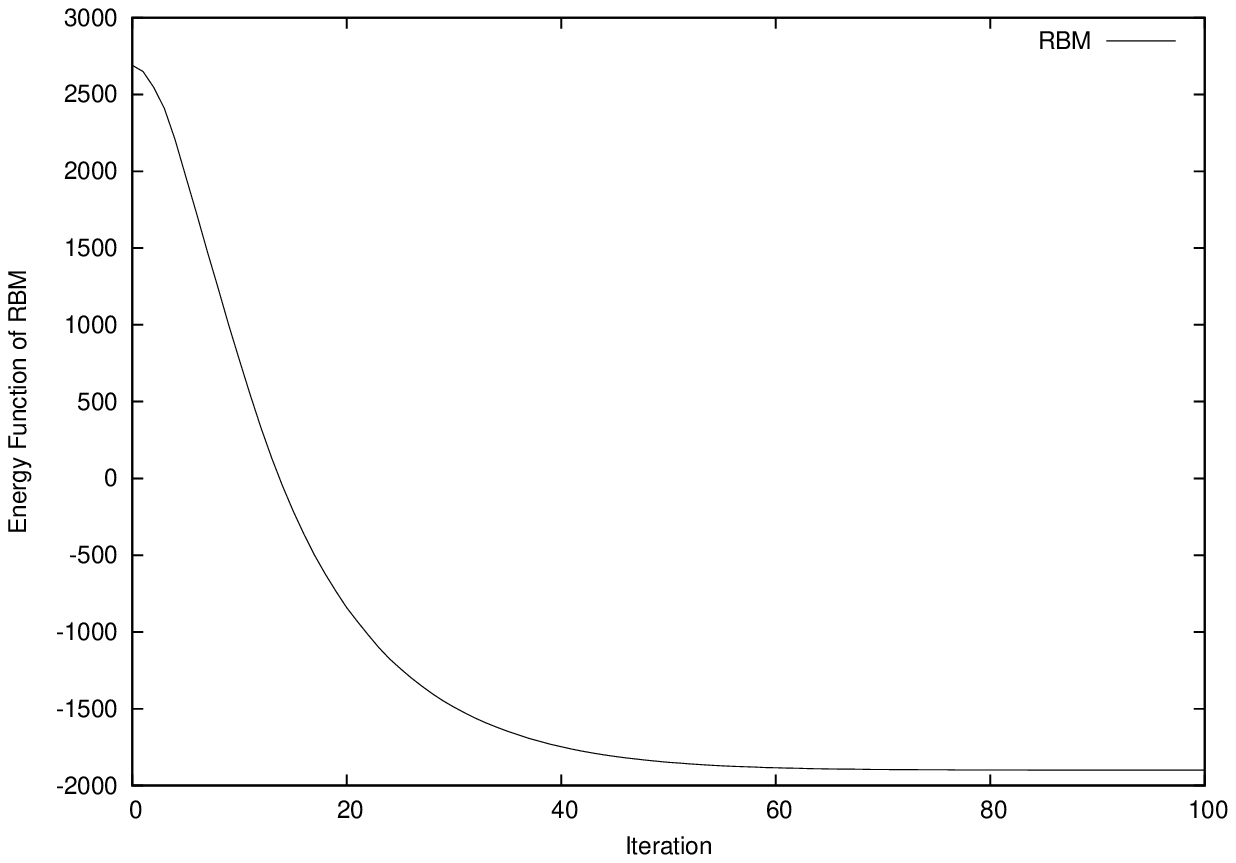}
\vspace{-3mm}
\caption{Energy Function of RBM}
\label{fig:result-rbm}
\end{center}
\end{figure}

Table \ref{tab:result-correct-ratio} shows the classification result for 3 kinds of images by CSAIM with perceptron and CSAIM with RBM. The upper line and the lower line show the result of training cases and test cases, respectively. Fig. \ref{fig:result-rbm} shows the experimental results of energy when a memory cell was generated by RBM.

\begin{figure}[tbp]
\begin{center}
\includegraphics[scale=0.4]{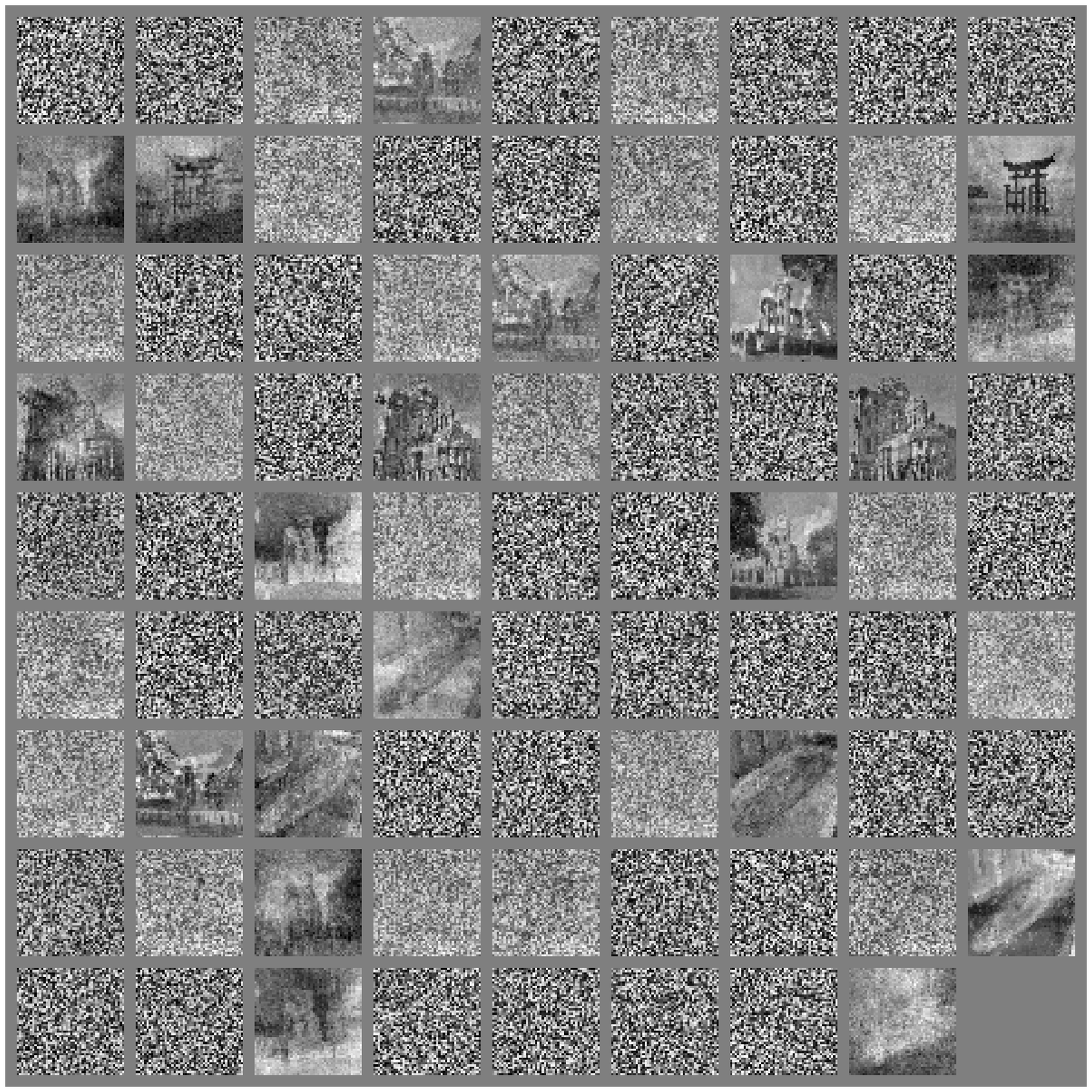}
\vspace{-3mm}
\caption{The visualization of weights learned by RBM}
\label{fig:weightspace}
\end{center}
\end{figure}

As the classification results, for data set of ``Torii'', the CSAIM with perceptron was able to classify them into correct label not only training cases but also test cases. But the previous method did not show good performance for the data set of ``Atomic Bomb Dome'' and ``Battleship Yamato''. The data set of them consists of unclear characteristics compared with ``Thorii''. For a point of view for human, we can detect the broken-building and battleship even if the object is captured from different angles. On the contrary, CSAIM with RBM was able to classify all the data sets not only training cases but also test cases. As a result, 3 memory cells were generated in the learning phase of CSAIM with RBM, Fig. \ref{fig:weightspace} shows the visual map of the weight space in a memory cell. The value of energy-function in RBM was decreased as shown in Fig. \ref{fig:result-rbm}. After 50 iterations, the energy became to fall into a small value, RBM was able to train faster than perceptron. The calculation speeds with CPU and GPU in RBM were 47.29 minutes and 25.22 minutes respectively, it was reduced by about half.

\begin{table*}[tbp]
\caption{Correct Ratio}
\vspace{-3mm}
\label{tab:result-correct-ratio}
\begin{center}
\begin{tabular}{r|r|r|r|r}
\hline \hline
                      & Correct ratio & Correct ratio(Torii in Miyajima) &  Correct ratio(Atomic Bomb Dome) & Correct ratio(Battleship Yamato)\\ \hline\hline
CSAIM                 &  33.3\%(8/24)  &    100.0\%(8/8) &    0.0\%(0/8) &   0.0\%(0/8) \\\cline{2-5}
(test)                &  33.3\%(2/6)    &   100.0\%(2/2) &    0.0\%(0/2) &   0.0\%(0/2) \\ \hline
CSAIM with RBM        &  83.3\%(20/24)  &  100.0\%(8/8)   &  100.0\%(8/8) &  50.0\%(4/8) \\\cline{2-5}
(test)                &  83.3\%(5/6)    &  100.0\%(2/2)   &  100.0\%(2/2) &  50.0\%(1/2) \\ 
\hline \hline
\end{tabular}
\end{center}
\end{table*}

\section{Conclusion}
\label{sec:Conclusion}
This paper presented the usage of Pylearn2 as one of machine learning tools for Deep Learning. Moreover, the generation method of Immunological Memory by using RBM was proposed to extract the features to classify the trained examples. As the experimental results, our proposed method showed the high classification capability for not only training cases but also test cases because some memory cells with characteristic pattern of images were generated by RBM. We will develop the method to find the optimum number of memory cells and the optimal set of values and the number of hidden units in the training phase of RBMs.

\vspace{-3mm}
\section*{Acknowledgment}
This work was supported by JSPS KAKENHI Grant Number 25330366 and Electric Technology Research Foundation of Chugoku.

\end{document}